\documentclass[10pt,twocolumn,letterpaper]{article}
\usepackage{iccv}

\definecolor{iccvblue}{rgb}{0.21,0.49,0.74}
\usepackage[pagebackref,breaklinks,colorlinks,allcolors=iccvblue]{hyperref}

\usepackage{enumitem, graphicx, multirow, amsfonts, amsmath, verbatim, algorithm, algorithmicx, algpseudocode, booktabs, array, arydshln, times, epsfig, amssymb, pifont, capt-of, wrapfig, lipsum, makecell, tabulary, etoolbox, xcolor}

\newcommand{\model}{UniVG}

\def\paperID{12852} \def\confName{ICCV} \def\confYear{2025}
\title{\model: A Generalist Diffusion Model for Unified Image Generation and Editing}
\author{Tsu-Jui Fu, Yusu Qian, Chen Chen, Wenze Hu, Zhe Gan, Yinfei Yang\\Apple}

\begin{document}

\twocolumn[{
\renewcommand\twocolumn[1][]{#1}
\maketitle
\begin{center}
    \includegraphics[width=.93\textwidth]{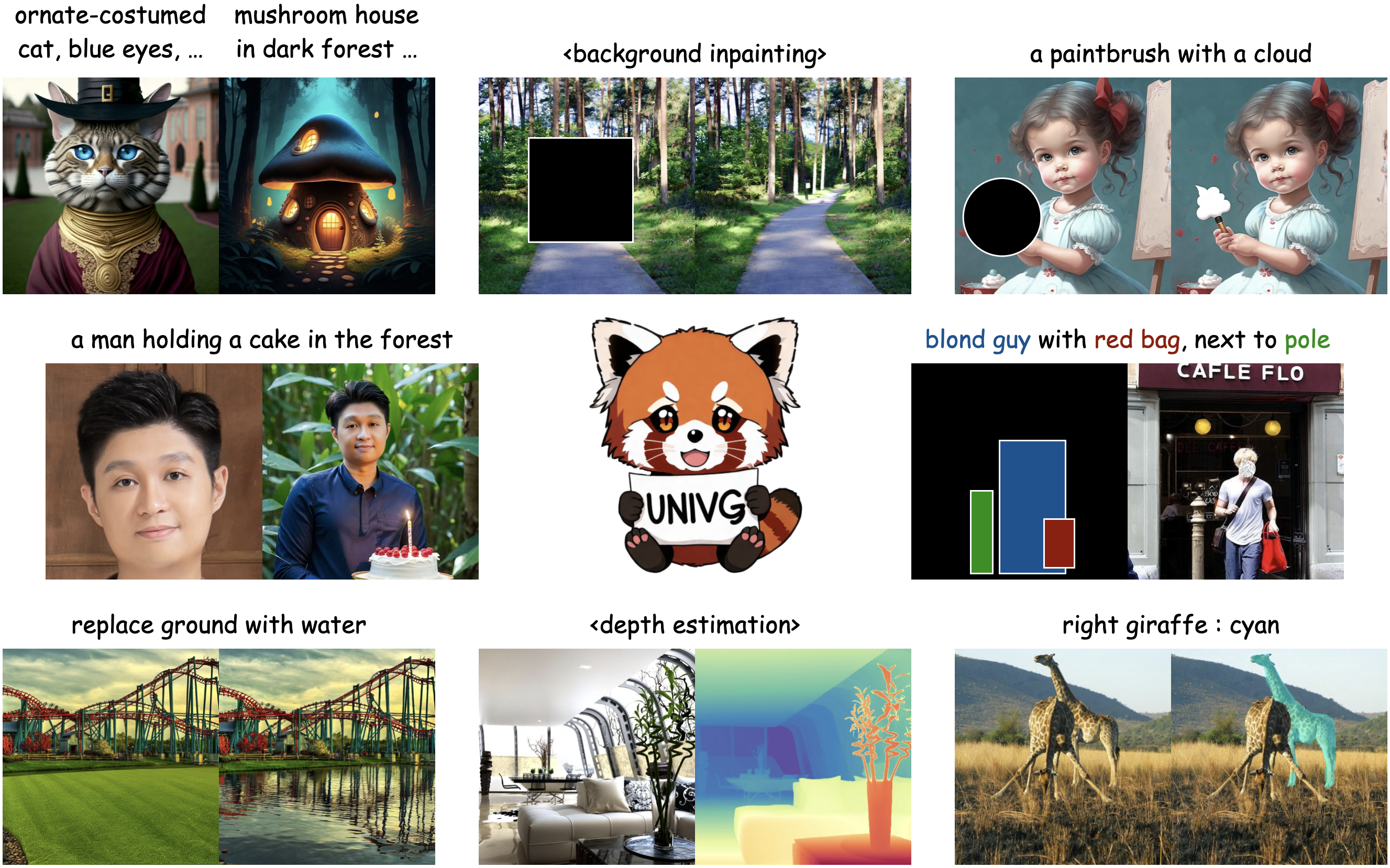}
    \vspace{-1.5ex}
    \captionof{figure}{We introduce \model, a single \emph{generalist} model that can support diverse image generation tasks, including text-to-image, inpainting, identity-preserving generation, layout-guided generation, instruction-based editing, depth estimation, and referring segmentation.}
    \label{fig:teaser}
    \vspace{1.5ex}
\end{center}
}]

\begin{abstract}
Text-to-Image (T2I) diffusion models have shown impressive results in generating visually compelling images following user prompts. Building on this, various methods further fine-tune the pre-trained T2I model for specific tasks. However, this requires separate model architectures, training designs, and multiple parameter sets to handle different tasks. In this paper, we introduce \model, a generalist diffusion model capable of supporting a diverse range of image generation tasks with a single set of weights. \model~treats multi-modal inputs as unified conditions to enable various downstream applications, ranging from T2I generation, inpainting, instruction-based editing, identity-preserving generation, and layout-guided generation, to depth estimation and referring segmentation. Through comprehensive empirical studies on data mixing and multi-task training, we provide detailed insights into the training processes and decisions that inform our final designs. For example, we show that T2I generation and other tasks, such as instruction-based editing, can coexist without performance trade-offs, while auxiliary tasks like depth estimation and referring segmentation enhance image editing. Notably, our model can even outperform some task-specific models on their respective benchmarks, marking a significant step towards a unified image generation model.
\end{abstract}
\section{Introduction}
Diffusion models, particularly those developed for text-to-image generation, have made significant strides. Models such as Stable Diffusion~\cite{rombach2022ldm,podell2023sdxl,esser2024sd3}, DALL-E~\cite{ramesh2022hierarchical}, and Imagen~\cite{baldridge2024imagen} have shown the capability to generate high quality, photorealistic images from text prompts. Meanwhile, various efforts have extended diffusion models to specialized tasks, leading to models such as InstructPix2Pix~\cite{brooks2023ins-p2p}, ControlNet~\cite{zhang2023adding}, and InstandID~\cite{wang2024instant-id}. However, the growing number of task-specific models has led to challenges in managing these systems efficiently and optimizing computational resources. A more scalable solution is a single, unified model capable of handling diverse image generation tasks, simplifying both development and deployment. This motivation has driven a growing interest in developing \emph{generalist} diffusion models in the community~\cite{xiao2024omni-gen,le2024one-diff}. 

In this paper, we present \model, a diffusion based model that unifies diverse image generation tasks within a single framework. Built on a minimally modified MM-DiT~\cite{esser2024sd3} architecture, \model~seamlessly integrates diverse types of inputs, including text prompts, masks, and existing images, and is able to adapt to different tasks by adjusting its inputs. Furthermore, external conditions (\emph{e.g.}, semantic maps or user-defined attributes) can be injected through embedding replacement to have further control. 

The concept of generalist diffusion models is not new, and has been explored in pioneering works such as OmniGen~\cite{xiao2024omni-gen} and OneDiffusion~\cite{le2024one-diff}. While these studies have demonstrated the feasibility of the approach and outlined high-level training procedures, the finer details of their execution remain unclear. Notably, both works lack clear ablation studies on optimal design choices for model training. To advance research in this area, we share our insights on building such models and focus on refining the best practices for developing a generalist diffusion model. Our investigation centers on three key aspects: ($i$) modeling, ($ii$) data recipe, and ($iii$) training strategy.

First, for \emph{modeling}, we adopt a \emph{minimalist} design, where the latent features of an input image are concatenated with the latent noise and the guided mask along the channel dimension, rather than the sequence dimension as in OmniGen~\cite{xiao2024omni-gen}. This minimalist design greatly improves training and inference efficiency compared to OmniGen~\cite{xiao2024omni-gen}, \emph{e.g.}, for instruction-based image editing (evidenced later in Section~\ref{sec:ablation-study}), allowing us to readily run large-scale experiments to investigate data recipe and training strategies. 

Second, for \emph{data recipe}, we curate a large-scale dataset encompassing diverse image generation tasks, ranging from text-to-image generation, inpainting, instruction-based editing, identity-preserving generation, layout-guided generation, to depth estimation and referring segmentation. Furthermore, we carefully examine the synergy between these tasks, an aspect unexplored in OmniGen~\cite{xiao2024omni-gen} and OneDiffusion~\cite{le2024one-diff}. For example, we find that instruction-based image editing does not compromise core text-to-image generation performance, while auxiliary tasks such as depth estimation and referring segmentation naturally enhance the performance of image editing.

Third, for \emph{training strategy}, instead of training on all data simultaneously, we adopt a progressive training approach. We first pretrain the model on large-scale text-to-image data, then gradually introduce instruction-based image editing and other image generation tasks. In the final stage, we mix in additional ID-preserving generation data to further fine-tune the model for this specific capability.

As illustrated in Fig.~\ref{fig:teaser}, by integrating all the insights presented in the paper, \model~achieves strong performance across all image generation tasks considered, demonstrating the advantages of a generalist model. Particularly, our \model~achieves a GenEval~\cite{ghosh2023gen-eval} score of 0.70, outperforming FLUX.1-dev~\cite{flux1} (with score 0.66), which is optimized solely for text-to-image generation.

Our main contributions are summarized as follow. 
($i$) We introduce \model, a generalist diffusion model capable of handling a wide range of image generation tasks without compromising core text-to-image generation performance.
($ii$) We present an in-depth study of data curation and training strategies, offering valuable insights for developing a unified image generation model.
($iii$) We achieve state-of-the-art performance compared to our two closest competitors, OmniGen~\cite{xiao2024omni-gen} and OneDiffusion~\cite{le2024one-diff}.

\begin{figure*}[t]
\centering
    \includegraphics[width=.89\linewidth]{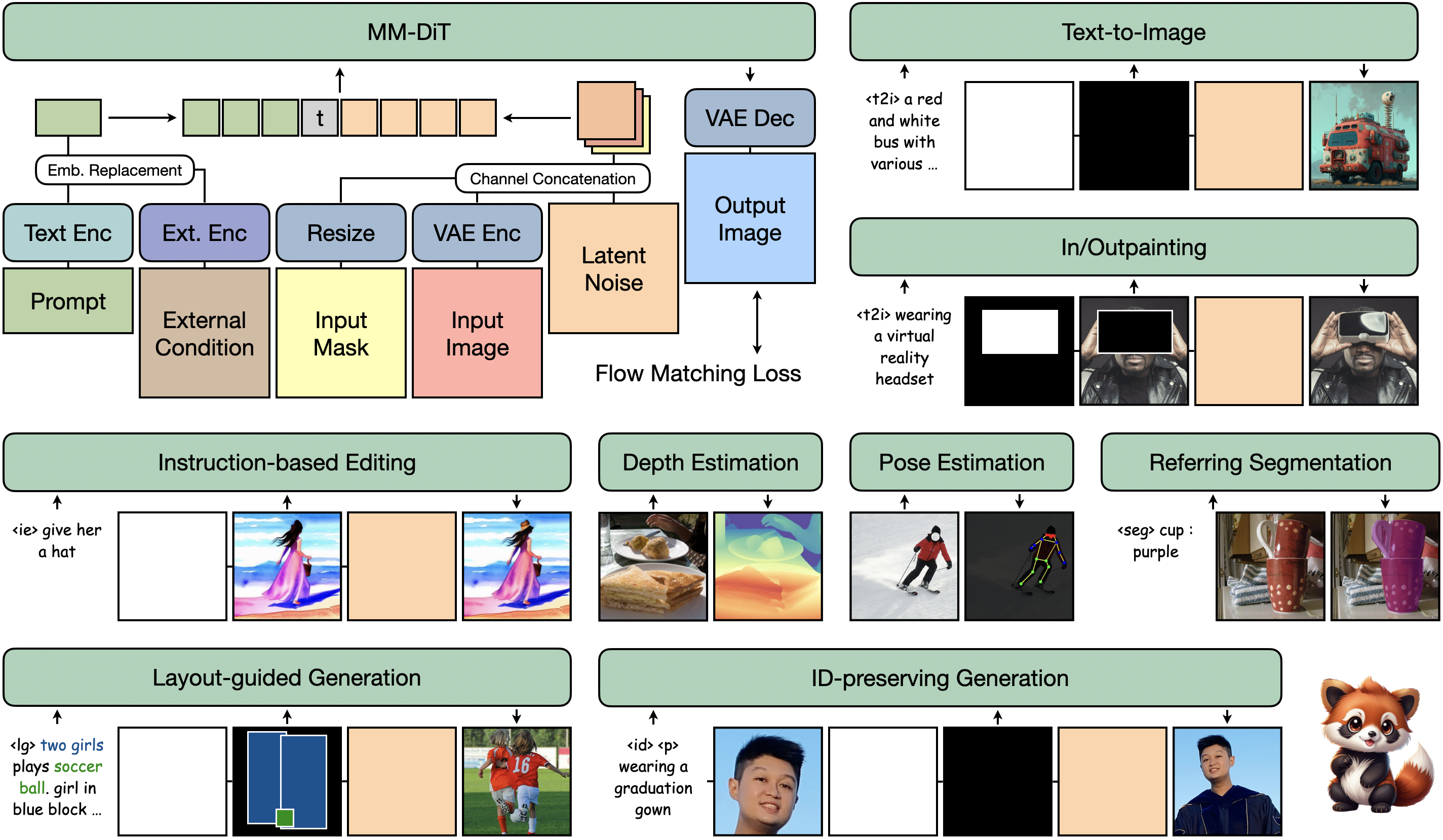}
    \vspace{-1.5ex}
    \caption{An overview of our \model. \model~contains a text encoder to extract prompt embeddings from the input text and an MM-DiT to perform cross-modal fusion for latent diffusion, where all visual guidance (latent noise, input image, and input mask) are concatenated along the channel dimension as a fix-length sequence for high efficiency. Additionally, an external condition can be injected through embedding replacement to have further control. Hence, a generalist \model~can support diverse tasks, such as text-to-image, in/outpainting, instruction-based editing, layout-guided generation, and ID-preserving generation. We also consider auxiliary tasks, including depth estimation, pose estimation, and referring segmentation, to enhance its visual scene perception.}
    \vspace{-3ex}
    \label{fig:architecture}
\end{figure*}

\section{Related Work}
\textbf{Latent Diffusion Models for Image Generation.}
Significant progress has been made in using diffusion models for image generation~\cite{song2019generative,ho2020ddpm,dhariwal2021diffusion}, making them the mainstream approach for text-to-image (T2I) tasks~\cite{nichol2021glide,baldridge2024imagen}. Building on latent-space image diffusion models such as Stable Diffusion~\cite{rombach2022ldm, podell2023sdxl}, recent work has increasingly adopted flow-based formulations~\cite{liu2023flow,tong2023improving,ma2024sit} and transformer-based architectures~\cite{peebles2022dit, bao2023all,hatamizadeh2024diffit,chen2023pixart}. The flow-based approach simplifies the image generation process, providing a more direct generation path that improves both model convergence speed and generation quality ~\cite{esser2024sd3}. In comparison with earlier U-Net architecture~\cite{ronneberger2015u-net,ho2020ddpm}, transformer-based models such as DiT~\cite{peebles2022dit} have a simpler design with fewer layer types and are more compatible with scale, benefiting from advancements in large language models. Both SD3~\cite{esser2024sd3} and Flux.1~\cite{flux1} integrate these advancements, establishing themselves as state-of-the-art open-source models.

Apart from improving the T2I performance, researchers have also explored using diffusion models for various other image generation applications, such as ($i$) fine-grained control~\cite{yang2023reco,li2023gligen,zhang2023adding,mou2024t2i}, ($ii$) instruction-based editing~\cite{brooks2023ins-p2p,li2023blip,fu2024mgie,zhang2023magic-brush}, ($iii$) personalized generation through conditioning on reference images~\cite{ye2023ip,wang2024instant-id,li2023photo-maker,guo2024pulid}, to name a few.

\vspace{1mm}
\noindent\textbf{Unified Diffusion Models.} 
Many works~\cite{qin2023unicontrol,xu2023versatile,zhao2023uni} have explored how to leverage diffusion models across different types of controls. However, these approaches are typically constrained to multiple image conditions and often require the design of complex adapters for each specific condition. Some other works~\cite{yang2022unitab,lu2023unified,lu2024unified,team2024chameleon}, such as TransFusion~\cite{zhou2024transfusion} and Show-o~\cite{xie2024show}, attempt to unify image understanding and generation. More recently, OmniGen~\cite{xiao2024omni-gen} and OneDiff~\cite{le2024one-diff} have introduced generalist diffusion models capable of handling a broad range of image generation tasks. In comparison, our approach offers a more thoroughly studied recipe for training such generalist models.

\section{Method}
\subsection{Background}
\noindent\textbf{Multi-modal Diffusion Transformer (MM-DiT).}
Different from the original cross-attention~\cite{rombach2022ldm,podell2023sdxl} via U-Net~\cite{ronneberger2015u-net}, MM-DiT leverages structured attention of DiT~\cite{peebles2022dit} to fuse features from multiple modalities and then perform the denoising process of a diffusion model~\cite{ho2020ddpm,song2021ddim}. Specifically, it concatenates all input as the single sequence and captures intricate cross-modal modeling to enhance both fidelity and controllability for multi-modal generation.

\vspace{1mm}
\noindent\textbf{Flow Matching.}
Classic denoising diffusion models, such as DDPM~\cite{ho2020ddpm} and DDIM~\cite{song2021ddim}, have shown promising ability in modeling complex data distribution (\textit{e.g.}, image generation). Instead of gradually adding Gaussian noises, flow matching~\cite{albergo2023flow,lipman2023flow-matching,liu2023flow} learns the continuous-time transformation. Specifically, a time-dependent vector field $u_t$ handles this transportation between noise and data, which is govern by ODE~\cite{chen2018ode} over $u$ from time 0 to time $t$. As $u$ is generally intractable~\cite{lipman2023flow-matching}, conditional flow matching (CFM)~\cite{lipman2023flow-matching,esser2024sd3} can learn a model $\mathcal{F}$ to imitate the ideal transformation:
\begin{equation}
    \mathcal{L}_\text{CFM} = \mathbb{E}~||\mathcal{F}(x_t, t~|~z) - u_t(x)||^2,
\end{equation}
where $x_t$ is the example $x$ at time $t$, and $z$ is the condition. Hence, the ODE solver does not require numerous discrete time steps~\cite{ho2020ddpm} and can adapt to the most efficient trajectory, leading to a more efficient sampling with fewer steps.

\subsection{\model}
Fig.~\ref{fig:architecture} illustrates the overview architecture of \model, which contains the text encoder to extract prompt embeddings $\{p\}$ for the input prompt $\mathcal{X}$ and MM-DiT $\mathcal{F}$ for diffusion modeling. Following latent diffusion learning~\cite{rombach2022ldm}, we apply Variational Autoencoder (VAE) for the input image $\mathcal{V}$, and the binary input mask $\mathcal{M}$ is resized accordingly. During training, the linear interpolation schedule~\cite{esser2024sd3} is applied over the output image $\mathcal{O}$ with Gaussian noise $\epsilon \sim \mathcal{N}(0, 1)$:
\begin{equation}
\begin{split}
    z &= \text{VAE}_\text{Enc}(\mathcal{O}), \\
    z_t &= t \cdot z + (1-t) \cdot \epsilon,
\end{split}
\end{equation}
where the target velocity field $u(z)=z-\epsilon$. We concatenate the latent noise $z_t$ with the visual inputs along the channel dimension into an equal-length sequence, which is the key to achieving high efficiency even considering multiple guidance and mitigating the context perception disruption ~\cite{mao2025ace++}. We unite the prompt embeddings and optimize $\mathcal{F}$ from the flow-matching loss:
\begin{equation} \label{eq:main}
\begin{split}
    d &= \lbrack z_t \oplus \text{VAE}_\text{Enc}(\mathcal{V}) \oplus \text{Resize}(\mathcal{M}) \rbrack, \\
    \mathcal{L} &= \mathbb{E}\left[||\mathcal{F}(\lbrack \{p\}, t, d \rbrack \rbrack)-u_t|| ^ 2\right].
\end{split}
\end{equation}

To add additional condition $\mathcal{C}$ for further control, we can utilize external encoder $\mathcal{H}$ that extracts domain-specific features $f=\mathcal{H}(\mathcal{C})$, which should have the same hidden dimension size to $\mathcal{F}$. We then inject this external condition by replacing the prompt embeddings of pre-designed placeholder tokens. For example, the facial features $f$ will substitute for ``\texttt{<p>}'' as our new prompt embeddings. By Eq.~\ref{eq:main}, MM-DiT can consider all guidance from prompt $\mathcal{X}$, input image $\mathcal{V}$, input mask $\mathcal{M}$, and external condition $\mathcal{C}$ for diverse control. Note that $\mathcal{C}$ is not limited to an image. Any format of guidance can be conditioned via its encoder. The length of $f$ is also flexible as long as having multiple placeholder tokens.

\vspace{1mm}
\noindent\textbf{Inference.} We follow classifier-free guidance (CFG)~\cite{ho2022cfg,brooks2023ins-p2p} during our \model~inference:
\begin{equation} \notag
\begin{split}
    \mathcal{F} &\implies \mathcal{F}(\varnothing, t, \{z_t, \varnothing, \varnothing\}) \\
    &+ \alpha_\mathcal{V} \cdot (\mathcal{F}(\varnothing, t, \{z_t, v, m) - \mathcal{F}(\varnothing, t, \{z_t, \varnothing, \varnothing\})) \\
    &+ \alpha_\mathcal{X} \cdot (\mathcal{F}(\{p\}, t, \{z_t, v, m) - \mathcal{F}(\varnothing, t, \{z_t, v, m\})),
\end{split}
\end{equation}
where $v$ is the latent features of $\mathcal{V}$, $m$ is the resized input mask of $\mathcal{M}$, and ($\alpha_\mathcal{V}$, $\alpha_\mathcal{X}$) is the guidance scale. After denoising back to $\hat{z}_0$, we utilize $\text{VAE}_\text{Dec}$ to get the actual image generation result.

\subsection{Multi-task Training}
To support various image generation applications, we consider diverse tasks and formulate each input format as follows for \model~multi-task training and inference (Fig.~\ref{fig:architecture}).

\vspace{1mm}
\noindent\textbf{Text-to-Image \& In/Outpainting.}
We prepend a special task token \texttt{<t2i>} for text-to-image, where the input image $\mathcal{V}$ is an empty (black) image, and the input mask $\mathcal{M}$ is all \texttt{True} (white), which means that we have to fill all regions in this generation. For in/outpainting, we reuse \texttt{<t2i>}, but $\mathcal{V}$ is an image with a black block, and $\mathcal{M}$ has a corresponding white block, which controls the model to paint the assigned region. During training, we randomly sample a region to mask out an image and treat its caption as the guided prompt~\cite{podell2023sdxl}. The complete image is the ideal output image $\mathcal{O}$. We further consider background in/outpainting, where the prompt is discarded as an empty string in this case.

\vspace{1mm}
\noindent\textbf{Instruction-based Editing.}
For instruction-based editing, $\mathcal{V}$ and $\mathcal{O}$ are the input and edited image, respectively. The prompt is the instruction with \texttt{<ie>} in front, with a blank $\mathcal{M}$ as all regions are editable.

\vspace{1mm}
\noindent\textbf{Auxiliary Tasks.}
To enhance the visual scene understanding of \model, we integrate depth estimation, pose estimation, and referring segmentation as our used auxiliary tasks. Rather than structured outputs, we follow OneDiff~\cite{le2024one-diff} and directly treat the visualization result of each task as $\mathcal{O}$ to learn via image generation. We utilize \texttt{<depth>} for depth estimation, \texttt{<pose>} for pose estimation, and \texttt{<seg>} for referring segmentation (with \texttt{target}:\texttt{color} in the prompt), where $\mathcal{V}$ is the input image, and $\mathcal{M}$ are all \texttt{True}.

\vspace{1mm}
\noindent\textbf{Layout-guided Generation.}
Regarding more fine-grained control, layout-guided generation requires the model to generate objects in assigned regions, where a given layout contains each bounding box of them. We visualize the layout as $\mathcal{V}$ and inject the object information into the prompt, such as ``\texttt{<lg>}~\textit{... girl in blue block. soccer ball in the green block.}'' In this way, \model~can have sufficient spatial guidance for layout-guided generation with an all-\texttt{True} $\mathcal{M}$. This highlights that, in our design, the input image is not necessarily limited to being visually similar to the output.

\vspace{1mm}
\noindent\textbf{ID-preserving Generation.}
We adopt the CLIP image encoder to extract facial embedding $f$ for an input face $\mathcal{C}$. We then replace its prompt embeddings $p$ of the placeholder token \texttt{<p>} and feed into MM-DiT. Therefore, \model~follows both input face and caption to perform ID customization. In detail, we apply the last layer of CLIP, followed by a two-layer MLP, as our used external encoder.

\begin{table}[t]
\small \centering \setlength{\tabcolsep}{2pt}
    \begin{tabular}{lr@{\hskip 0.1in}|@{\hskip 0.1in}lr}
        \toprule
        \textbf{Task} & \textbf{Ratio} & \textbf{Task} & \textbf{Ratio} \\
        \midrule
        Text-to-Image & 28\% & Instruction-based Editing & 47\% \\
        Inpainting & 10\% & Auxiliary Tasks & 3\% \\
        Outpainting & 10\% & Layout-guided Generation & 2\% \\
        \bottomrule
    \end{tabular}
    \vspace{-2ex}
    \caption{The used mixture for \model~multi-task training.}
    \label{table:multi-task-mixture}
    \vspace{-3ex}
\end{table}

\vspace{1mm}
\noindent\textbf{Training Recipe.}
We present the used multi-stage training recipe of \model~based on our empirical observations:
\begin{itemize}
    \item \textbf{Stage I (foundation training)}: We train MM-DiT from scratch on text-to-image with lr=1e-4 and batch\_size=512 for 400K steps;
    \item \textbf{Stage II (multi-task training)}: We have in/outpainting, instruction-based editing, auxiliary tasks, layout-guided generation along with text-to-image for multi-task training. The detailed mixture is shown in Sec.~\ref{sec:dataset}, where we also adopt lr=1e-4 and batch\_size=512 for 400K steps;
    \item \textbf{Stage III (further finetuning)}: After finding the catastrophic forgetting issue if involving ID-preserving generation at Stage II, we instead train this ID-customization task with all other multi-task data in a 1:1 ratio afterward. The used external image encoder is also trained with MM-DiT for lr=2e-5, batch\_size=512, and 40K steps.
\end{itemize}
We conduct a comprehensive ablation study in Sec.~\ref{sec:ablation-study}.

\section{Experiments}

\begin{table}[t]
\small \centering \setlength{\tabcolsep}{2pt}
    \begin{tabular}{lrcccc}
        \toprule
        \textbf{Method} & \textbf{\#Param} & \textbf{GenEval}$\uparrow$ & \textbf{CompBench}$\uparrow$ & \textbf{DSG}$\uparrow$ & \textbf{HPSv2}$\uparrow$ \\
        \midrule
        SDXL & 2.6B & 0.55 & 0.42 & 0.72 & 27.7 \\
        FLUX.1 & 12.0B & 0.66 & 0.47 & 0.73 & 29.2 \\
        SD3 & 8.0B & 0.71 & 0.49 & 0.76 & 28.9 \\
        \midrule
        OneDiff & 2.8B & 0.65 & 0.44 & \underline{0.68} & 27.5 \\
        OmniGen & 3.8B & \underline{0.70} & \underline{0.46} & 0.66 & \underline{27.7} \\
        \model & 3.7B & \textbf{0.70} & \textbf{0.48} & \textbf{0.75} & \textbf{28.2} \\
        \bottomrule
    \end{tabular}
    \vspace{-2ex}
    \caption{Results of text-to-image generation on GenEval~\cite{ghosh2023gen-eval}, T2I-CompBench~\cite{huang2023comp-bench}, DSG~\cite{cho2024dsg}, and HPSv2~\cite{wu2023hpsv2}.}
    \label{table:text-to-image}
    \vspace{-3ex}
\end{table}

\subsection{Experimental Setup}
\noindent\textbf{Datasets.} \label{sec:dataset}
We construct a dataset collection to build a generalist model that supports diverse tasks. For text-to-image, we have internal 2B text-image pairs, JourneyDB-4M~\cite{sun2023jdb}, and DALLE3-1M~\cite{egan2024dalle3-1m}. We consider two scenarios of image in/outpainting: text-guided and background. We utilize our text-image pairs for text-guided in/outpainting; we involve our internal 5M scene images, OSV-5M~\cite{astruc2024osv-5m}, and Places365-1M~\cite{zhou2016places365} for background in/outpainting.

We incorporate open-source datasets, including IPr2Pr-1M~\cite{brooks2023ins-p2p}, UltraEdit-4M~\cite{zhao2024ultra-edit}, SeedEdit-3M~\cite{ge2024seed-edit}, OmniEdit-1M~\cite{wei2024omni-edit}, and StyleBooth-11K~\cite{han2024style-booth} for our instruction-based editing. All of them contain triplets of (input image, instruction, output image). Most of the image pairs are synthesized from Prompt-to-Prompt~\cite{hertz2023prompt-to-prompt} or inpainting. In our auxiliary tasks, we consider COCO-118K~\cite{lin2014coco}, KITTI-7K~\cite{geiger2012kitti}, and Hypersim-75K~\cite{roberts2021hypersim} for depth estimation, COCO-27K~\cite{lin2014coco} for pose estimation, and COCO-213K~\cite{lin2014coco}, RefCOCO~\cite{yu2016ref-coco}, and PhraseCut-298K~\cite{wu2020phrase-cut} for referring segmentation.

To support more guidance, we include Flickr-148K~\cite{young2014flickr30k} and SBU-840K~\cite{ordonez2021sbu} for layout-guided generation and follow the pre-processing in GLIGEN~\cite{li2023gligen} to acquire each object and corresponding bounding box. We collect our internal 603K images with clear human faces for ID-preserving generation, where the cropped face is the input ID, and the original image is the target output. We utilize the caption of the whole image as the input prompt. The used mixture for \model~multi-task training (stage II) is presented in Table~\ref{table:multi-task-mixture}. At stage III, we set a 1:1 ratio between ID-preserving generation and all other tasks to further learn ID customization and keep the original ability in generation and editing.

\vspace{1mm}
\noindent\textbf{Implementation Details.}
We treat internal CLIP-bigG~\cite{lai2024ve-clip} as the text encoder, and \model~contains 38 layers of MM-DiT with a hidden dimension size of 2432 and 38 attention heads, leading to a total of 3.7B model. We apply an internal 8-channel VAE to extract the latent features of an image for diffusion modeling. For ID-preserving generation, we have the CLIP image encoder as the external encoder for a face. We follow the recipe to train \model~using Adafator~\cite{shazeer2018adafactor} on 512-v5p TPUs. During inference, we set the guidance scale ($\alpha_\mathcal{X}$, $\alpha_\mathcal{V}$) to (4.0, 1.5). All implementations are done using the AXLearn framework\footnote{AXLearn: \url{https://github.com/apple/axlearn}}.

\subsection{Evaluation Results}

\begin{figure*}[t]
\centering
    \includegraphics[width=.89\linewidth]{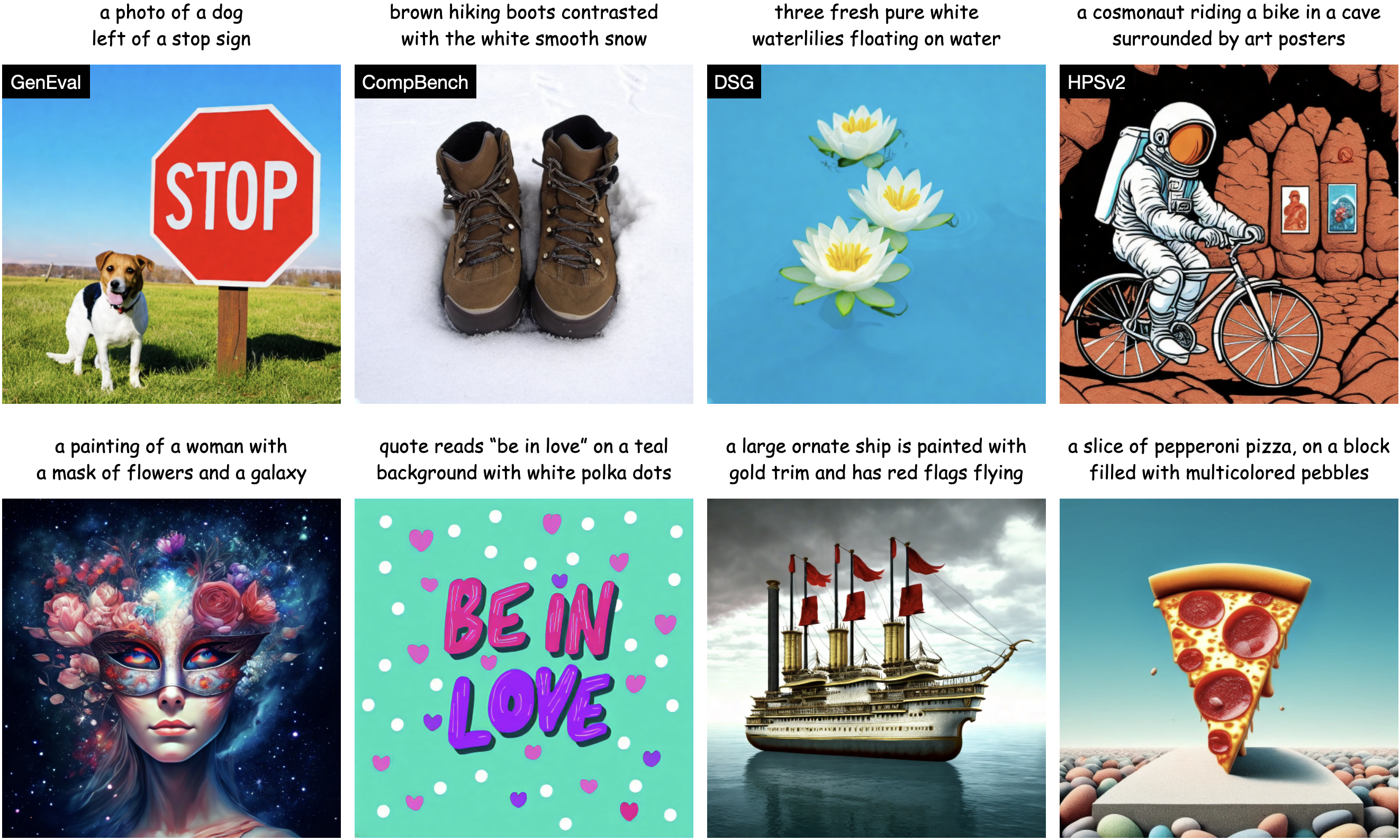}
    \vspace{-1.5ex}
    \caption{Qualitative examples of text-to-image generation. Note that we simplify the prompt for better presentation.}
    \vspace{-3ex}
    \label{fig:text-to-image}
\end{figure*}

\noindent\textbf{Text-to-Image Generation.}
As there are many aspects to study the quality of text-to-image, we adopt GenEval~\cite{ghosh2023gen-eval} and T2I-CompBench~\cite{huang2023comp-bench} for the compositionality, such as object, texture, color, and relative position. We also include DSG~\cite{cho2024dsg}, which utilizes visual-question answering to verify whether the generated image aligns with the prompt. From HPSv2~\cite{wu2023hpsv2}, we investigate the visual quality with respect to the pre-trained human preference score. We consider contemporary unified models, OneDiff~\cite{le2024one-diff} and OmniGen~\cite{xiao2024omni-gen}, as our main baselines. In addition, we also treat SDXL~\cite{podell2023sdxl}, FLUX.1~\cite{flux1}, and SD3~\cite{esser2024sd3} as specific text-to-image methods for the comparison. Table~\ref{table:text-to-image} shows that our \model~surpasses those unified baselines on both compositionality and semantic across all benchmarks. These results support that we can precisely follow the prompt yet perform visually appealing text-to-image. Note that \model~has even fewer parameters than OmniGen, which highlights the advantage of our model design and multi-task training. Furthermore, we achieve competitive performance to the task-specific SD3, which contains twice more parameters than ours. This also encourages the potential of the generalist model, where we can maintain the capability to handle various tasks well.

\begin{table}[t]
\small \centering \setlength{\tabcolsep}{2pt}
    \begin{tabular}{lccccc}
        \toprule
        ~ & \multicolumn{2}{c}{\textbf{MagicBrush}} & ~ & \multicolumn{2}{c}{\textbf{EmuEdit}} \\
        \cmidrule{2-3} \cmidrule{5-6}
        \textbf{Method} & CLIP-T$\uparrow$ & CLIP-I$\uparrow$ & ~ & CLIP-T$\uparrow$ & CLIP-I$\uparrow$ \\
        \midrule
        InsP2P & 24.5 & 83.7 & ~ & 21.9 & 83.4 \\
        MGIE & 26.4 & 84.6 & ~ & 22.4 & 84.2 \\
        EmuEdit & 26.1 & 89.7 & ~ & 23.1 & 85.9 \\
        \midrule
        OneDiff & 24.8 & \textbf{88.5} & ~ & 22.0 & \textbf{85.5} \\
        OmniGen & \underline{25.8} & 86.3 & ~ & \underline{23.1} & 82.9 \\
        \model & \textbf{29.5} & \underline{86.3} & ~ & \textbf{25.9} & \underline{84.7} \\
        \bottomrule
    \end{tabular}
    \vspace{-2ex}
    \caption{Results of instruction-based editing on MagicBrush~\cite{zhang2023magic-brush} and EmuEdit~\cite{sheynin2023emu-edit}.}
    \label{table:instruction-based-editing}
    \vspace{-3ex}
\end{table}

\begin{figure*}[t]
\centering
    \includegraphics[width=.782\linewidth]{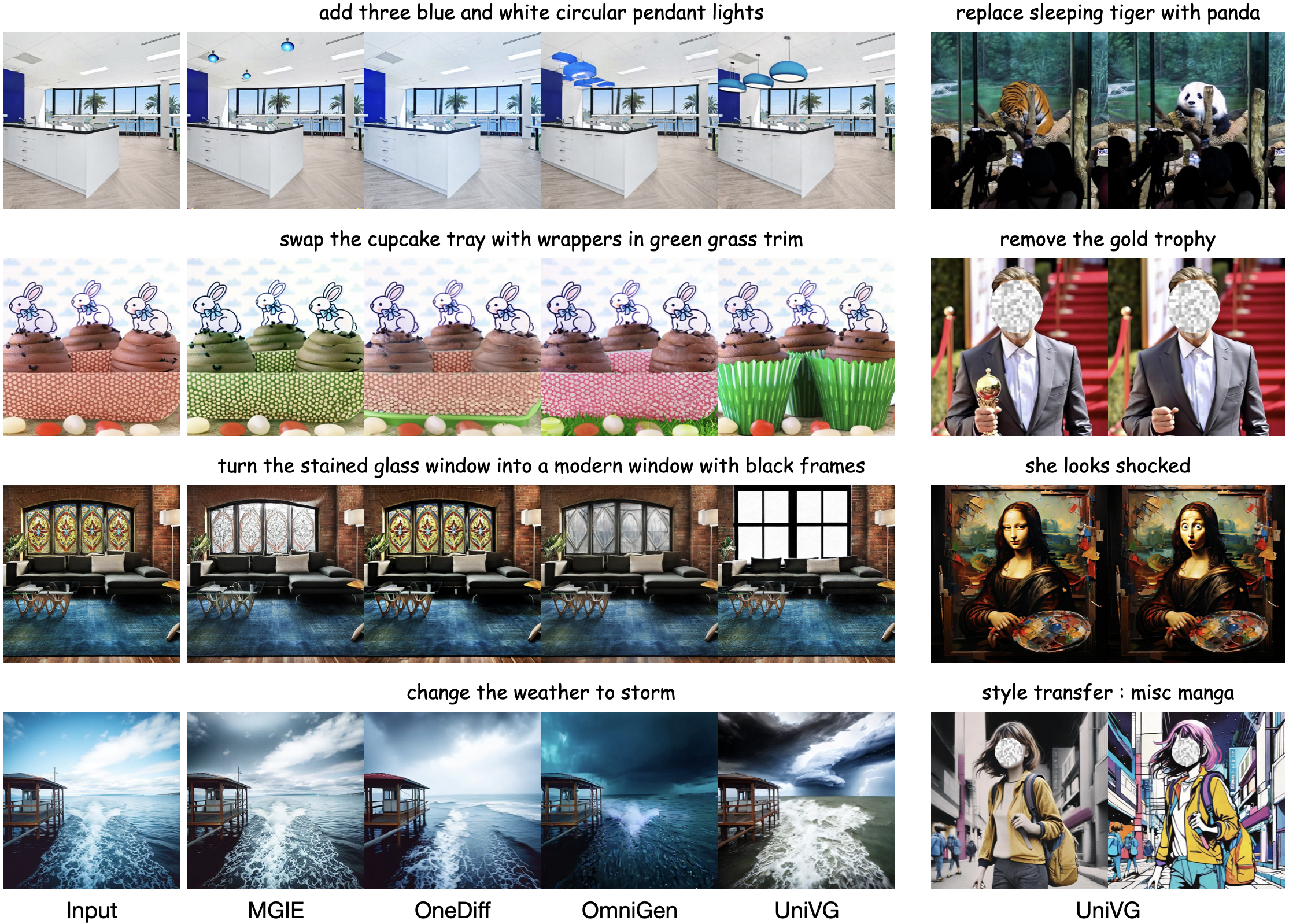}
    \vspace{-1.5ex}
    \caption{Qualitative comparisons of instruction-based editing.}
    \vspace{-1ex}
    \label{fig:instruction-based-editing}
\end{figure*}

\vspace{1mm}
\noindent\textbf{Instruction-based Editing.}
To evaluate instruction-based editing, we consider the testing set of MagicBrush~\cite{zhang2023magic-brush} and EmuEdit~\cite{sheynin2023emu-edit}. We use CLIP-T~\cite{hessel2021clip-score} between the target caption and the edited image for text-visual alignment. We also apply CLIP-I to investigate the input preservation. We have InsP2P~\cite{brooks2023ins-p2p}, MGIE~\cite{fu2024mgie}, EmuEdit~\cite{song2021ddim} as specific models for editing. Table~\ref{table:instruction-based-editing} comparing our \model~with baselines. Surprisingly, we even achieve significantly higher CLIP-T than task-specific methods, which shows our strong ability in instruction following to modify an existing image. In comparison with the unified OmniGen, \model~has comprehensive advantages in CLIP-T and CLIP-I, which are the two trade-off goals of instruction-based editing. Though OneDiff has higher CLIP-I, its modification is usually limited and results in low CLIP-T that cannot meet the editing expectation. For example, in Fig.~\ref{fig:instruction-based-editing}, OneDiff fails to add circular lights to the ceiling. MGIE and OmniGen are attempting, but the results do not look visually appealing. In contrast, \model~follows the same visual flow for editing. Moreover, we are the only model that can achieve complex modifications, such as replacing with ``\textit{green grass wrapper}'' or ``\textit{black frames}''. Our \model~also supports diverse purposes, including removal, facial emotion, and overall artistic stylization. These qualitative results illustrate the strength of \model~for universal instruction-based image editing.

\begin{figure}[t]
\begin{minipage}{.51\linewidth}
\small \centering \setlength{\tabcolsep}{2pt}
    \begin{tabular}{lcc}
        \toprule
        ~ & \multicolumn{2}{c}{\textbf{Unsplash-50}} \\
        \cmidrule{2-3}
        \textbf{Method} & ID$\uparrow$ & CLIP-T$\uparrow$ \\
        \midrule
        PhotoMaker & 0.193 & 27.4 \\
        InstantID & 0.648 & 26.4 \\
        PuLID & 0.654 & 31.2 \\
        \midrule
        OneDiff & 0.283 & 26.8 \\
        OmniGen & \underline{0.294} & \underline{27.1} \\
        \model & \textbf{0.329} & \textbf{28.1} \\
        \bottomrule
    \end{tabular}
    \vspace{-2ex}
    \captionof{table}{Results of ID-preserving generation on Unsplash-50~\cite{gal2024unsplash-50}.}
    \label{table:id-preserving-generation}
    \vspace{-3ex}
\end{minipage}~~
\begin{minipage}{.44\linewidth}
\vspace{8.5ex}
\small \centering \setlength{\tabcolsep}{2pt}
    \begin{tabular}{ccc}
        \toprule
        ~ & \multicolumn{2}{c}{\textbf{Unsplash-50}} \\
        \cmidrule{2-3}
        \textbf{3-Stage} & ID$\uparrow$ & CLIP-T$\uparrow$ \\
        \midrule
        \ding{55} & 0.245 & \textbf{28.3} \\
        \ding{51} & \textbf{0.328} & 28.1 \\
        \bottomrule
    \end{tabular}
    \vspace{-2ex}
    \captionof{table}{Training all at once results in worse performance for ID preservation.}
    \label{table:3-stage}
    \vspace{-3ex}
\end{minipage}
\end{figure}

\vspace{1mm}
\noindent\textbf{ID-preserving Generation.}
We adopt Unsplash-50~\cite{gal2024unsplash-50} to evaluate ID-preserving generation, which provides human faces and descriptions to make the model generate personalized images. We then follow CurricularFace~\cite{huang2020curricular-face} to calculate the facial embeddings for ID similarity and CLIP-T for the prompt-image score. We treat PhotoMaker~\cite{li2023photo-maker}, InstantID~\cite{wang2024instant-id}, and PuLID~\cite{guo2024pulid} for the task-specific models. In Table~\ref{table:id-preserving-generation}, \model~again outperforms the unified baselines with notable improvements in face consistency, which highlights the effectiveness of our design to inject flexible guidance and control the further generation. Moreover, a higher CLIP-T demonstrates that we also lead to a superior prompt following for ID customization. Compared to task-specific methods, InstantID and PuLID rely on the pre-trained face encoder to bring strong ID preservation. Nevertheless, their generated faces are significantly limited to inputs and cannot support complex manipulations~\cite{le2024one-diff}. Table~\ref{table:3-stage} highlights the cruciality of our carefully designed multi-stage training, where training all at once results in catastrophic forgetting for ID preservation (notably lower ID similarity).

\begin{table*}[t]
\small \centering \setlength{\tabcolsep}{2pt}
    \begin{tabular}{cccccccccccccccccccc}
        \toprule
        ~ & ~ & \multicolumn{4}{c}{\textbf{Task}} & ~ & ~ & ~ & ~ & ~ & ~ & \multicolumn{2}{c}{\textbf{MagicBrush}} & ~ & \multicolumn{2}{c}{\textbf{EmuEdit}} & ~ & \multicolumn{2}{c}{\textbf{Unsplash-50}} \\
        \cmidrule{3-6} \cmidrule{13-14} \cmidrule{16-17} \cmidrule{19-20}
        ~ & ~ & \ding{171} & \ding{170} & \ding{169} & \ding{168} & ~ & \textbf{GenEval}$\uparrow$ & \textbf{CompBench}$\uparrow$ & \textbf{DSG}$\uparrow$ & \textbf{HPSv2}$\uparrow$ & ~ & CLIP-T$\uparrow$ & CLIP-I$\uparrow$ & ~ & CLIP-T$\uparrow$ & CLIP-I$\uparrow$ & ~ & ID$\uparrow$ & CLIP-T$\uparrow$ \\
        \midrule
        (a) & ~ & \ding{51} & \ding{55} & \ding{55} & \ding{55} & ~ & \textbf{0.71} & \textbf{0.49} & \textbf{0.76} & \textbf{28.4} & ~ & \multicolumn{2}{c}{-} & ~ & \multicolumn{2}{c}{-} & ~ & \multicolumn{2}{c}{-} \\
        (b) & ~ & \ding{51} & \ding{51} & \ding{55} & \ding{55} & ~ & \underline{0.70} & \underline{0.48} & \underline{0.76} & \underline{28.3} & ~ & 28.3 & \underline{88.0} & ~ & 25.2 & \textbf{86.1} & ~ & \multicolumn{2}{c}{-} \\
        (c) & ~ & \ding{51} & \ding{51} & \ding{55} & \ding{51} & ~ & 0.70 & 0.48 & 0.75 & 28.2 & ~ & 26.9 & \textbf{88.2} & ~ & 24.9 & \underline{85.4} & ~ & 0.327 & \textbf{28.9} \\
        (d) & ~ & \ding{51} & \ding{51} & \ding{51} & \ding{55} & ~ & 0.70 & 0.48 & 0.75 & 28.1 & ~ & \textbf{29.8} & 87.4 & ~ & \textbf{26.2} & 84.8 & ~ & \multicolumn{2}{c}{-} \\
        (e) & ~ & \ding{51} & \ding{51} & \ding{51} & \ding{51} & ~ & 0.70 & 0.48 & 0.75 & 28.2 & ~ & \underline{29.5} & 86.3 & ~ & \underline{25.9} & 84.7 & ~ & \textbf{0.329} & 28.1 \\
        \bottomrule
    \end{tabular}
    \vspace{-2ex}
    \caption{Ablation study of multi-task training. \ding{171} Text-to-Image Generation and In/outpainting; \ding{170} Instruction-based Editing; \ding{169} Auxiliary Tasks and Layout-guided Generation; \ding{168} ID-preserving Generation. Our receipe: Stage I (\ding{171}); Stage II (\ding{171}+\ding{170}+\ding{169}); Stage III (\ding{171}+\ding{170}+\ding{169}+\ding{168}).}
    \label{table:ablation-study}
    \vspace{-2ex}
\end{table*}

\begin{table}[t]
\small \centering \setlength{\tabcolsep}{2pt}
    \begin{tabular}{lrrrcrr}
        \toprule
        ~ & ~ & \multicolumn{2}{c}{\textbf{Text-to-Image}} & ~ & \multicolumn{2}{c}{\textbf{Editing}} \\
        \cmidrule{3-4} \cmidrule{6-7}
        \textbf{Method} & \textbf{\#Param} & Time & GPU & ~ & Time & GPU \\
        \midrule
        OneDiff & 2.8B & 6.3 & 8151 & ~ & 10.8 & 9155 \\
        OmniGen & 3.8B & 9.3 & 8813 & ~ & 36.8 & 11895 \\
        \model~& 3.7B & 10.4 & 8849 & ~ & 10.4 & 8849 \\
        \bottomrule
    \end{tabular}
    \vspace{-2ex}
    \caption{Results of efficiency comparisons on time (sec) and GPU cost (MB) during inference (512$^2$).}
    \label{table:efficiency}
    \vspace{-3ex}
\end{table}

\subsection{Ablation Study} \label{sec:ablation-study}
Table.~\ref{table:ablation-study} presents the ablation study of our multi-task training. Compared to row (a), row (b) shows strong instruction-based editing, yet maintains a competitive performance in text-to-image generation. This points out that learning both together will not hurt either but can enable these two abilities in a single model. Row (c) then trains on ID-preserving generation. However, there is a notable drop in editing (\textit{e.g.}, CLIP-T from 28.3 to 26.9 on MagicBrush) as the model is placing more on ID customization. To overcome this issue, we involve the auxiliary tasks to enhance the visual understanding of \model. Row (d) gains further improvements in editing over row (b), which highlights the usage of auxiliary tasks. This time, even with ID-preserving generation, row (e) strikes the best balance with all favorable performance.

\begin{figure}[t]
\centering
    \includegraphics[width=.89\linewidth]{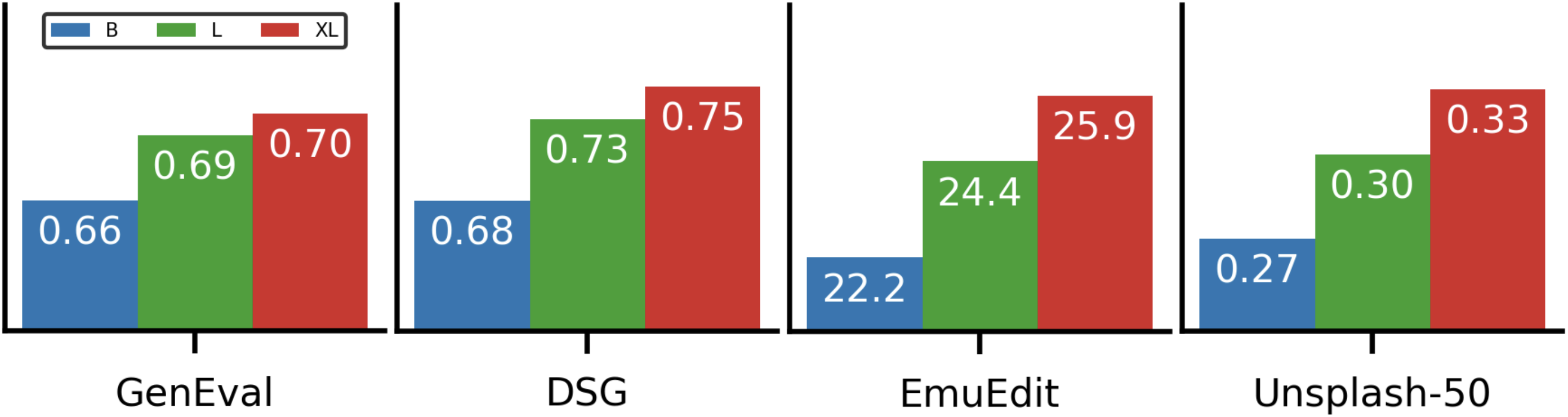}
    \vspace{-1.5ex}
    \caption{Resuling of model scaling with B (416M), L (1.8B), and XL (3.7B); CLIP-T for EmuEdit and ID for Unsplash-50.}
    \vspace{-3ex}
    \label{fig:scale}
\end{figure}

\vspace{1mm}
\noindent\textbf{Inference Efficiency.}
In addition to generation quality, we also investigate the inference time and GPU cost on a single NVIDIA A100 GPU in Table.~\ref{table:efficiency}. This comparison is done with image resolution 512$^2$ and model precision BFloat16. Both OneDiff and OmniGen are compelled to concatenate additional images with the noise sequence, which significantly increases the computation overhead of MM-DiT, resulting in a notable degradation for editing (\textit{e.g.}, OmniGen requires 36+ seconds). On the other hand, our \model~considers the latent noise, input image, and mask all along with the channel dimension, which can maintain the same total sequence length for both generation and editing. Therefore, we even bring a better efficiency to OneDiff in editing, yet with a larger model. These observations also highlight that \model~not only leads to superior image generation but also consistently high efficiency across diverse tasks.

\vspace{1mm}
\noindent\textbf{Model Scaling.}
We study the model scaling performance of unified image generation. We consider three model sizes, including B (416M with 18 layers), L (1.8B with 30 layers), and XL (3.7B with 38 layers). Fig.~\ref{fig:scale} illustrates that as scaling up, the performance keeps improving and generalizes to generation and editing tasks. This encourages the immense potential for more powerful and capable future models.

\begin{figure}[t]
\centering
    \includegraphics[width=.89\linewidth]{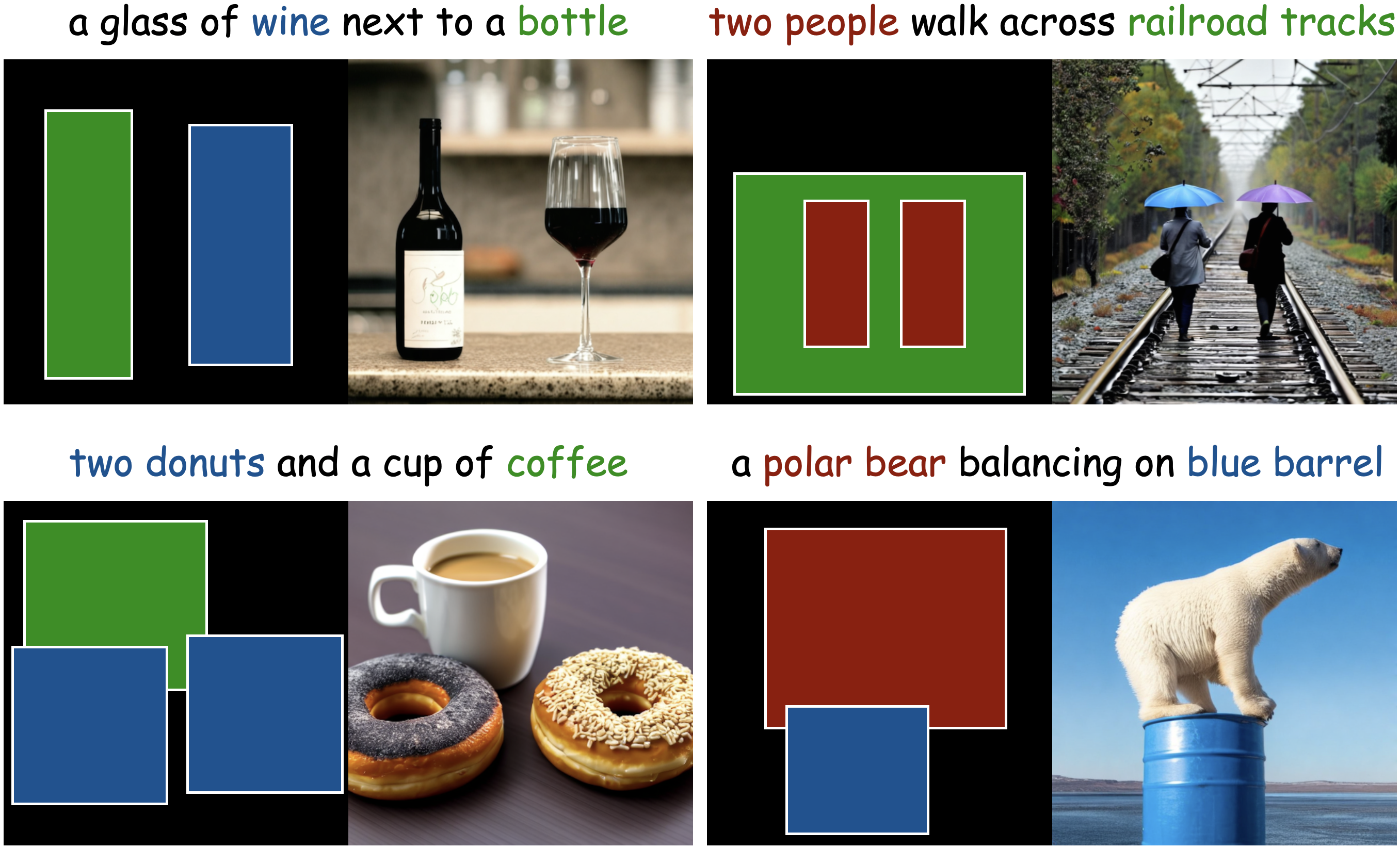}
    \vspace{-1.5ex}
    \caption{Qualitative examples of layout-guided generation, where the colors of blocks are aligned with the objects in the prompt.}
    \vspace{-1ex}
    \label{fig:layout-guided-generation}
\end{figure}

\begin{figure}[t]
\centering
    \includegraphics[width=.89\linewidth]{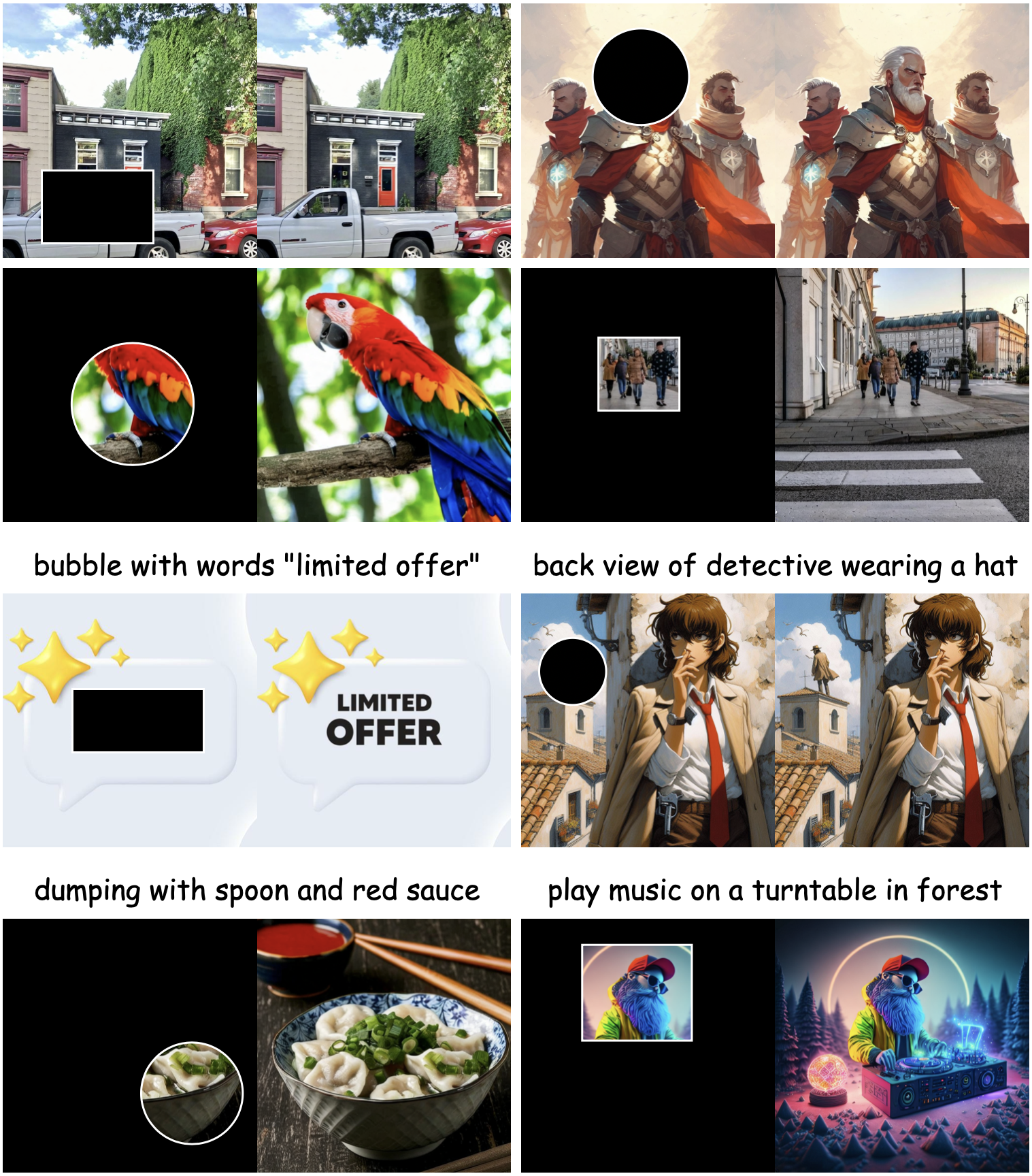}
    \vspace{-1.5ex}
    \caption{Qualitative examples of background in/outpainting and text-guided in/outpainting.}
    \vspace{-3ex}
    \label{fig:painting}
\end{figure}

\vspace{1mm}
\noindent\textbf{More Visualization Results.}
Fig.~\ref{fig:layout-guided-generation} presents layout-guided generation, where the actual prompt is joined with the layout (\textit{e.g.}, ``\textit{a glass of wine next to a bottle. wine in the blue block. bottle in the green block.}''). \model~follows the spatial guidance of each object to generate the image. We can also deal with counting (\textit{e.g.}, \textit{two donuts}) and present them in the assigned positions.

Fig.~\ref{fig:painting} shows the results of in/outpainting. Though without prompts, our \model~still recovers the missing regions (\textit{e.g.}, car and human face). We can even imagine the overall visual scene from just a small block and perform outpainting to expand it as a reasonable image (\textit{e.g.}, parrot and street view). \model~further controls this through the text, where we can inpaint a specific draw text (\textit{e.g.}, ``\textit{limited offer}'') or object (\textit{e.g.}, \textit{detective}). Similarly, we guide and outpaint the whole image to support creative image completion. 

\begin{figure}[t]
\centering
    \includegraphics[width=.89\linewidth]{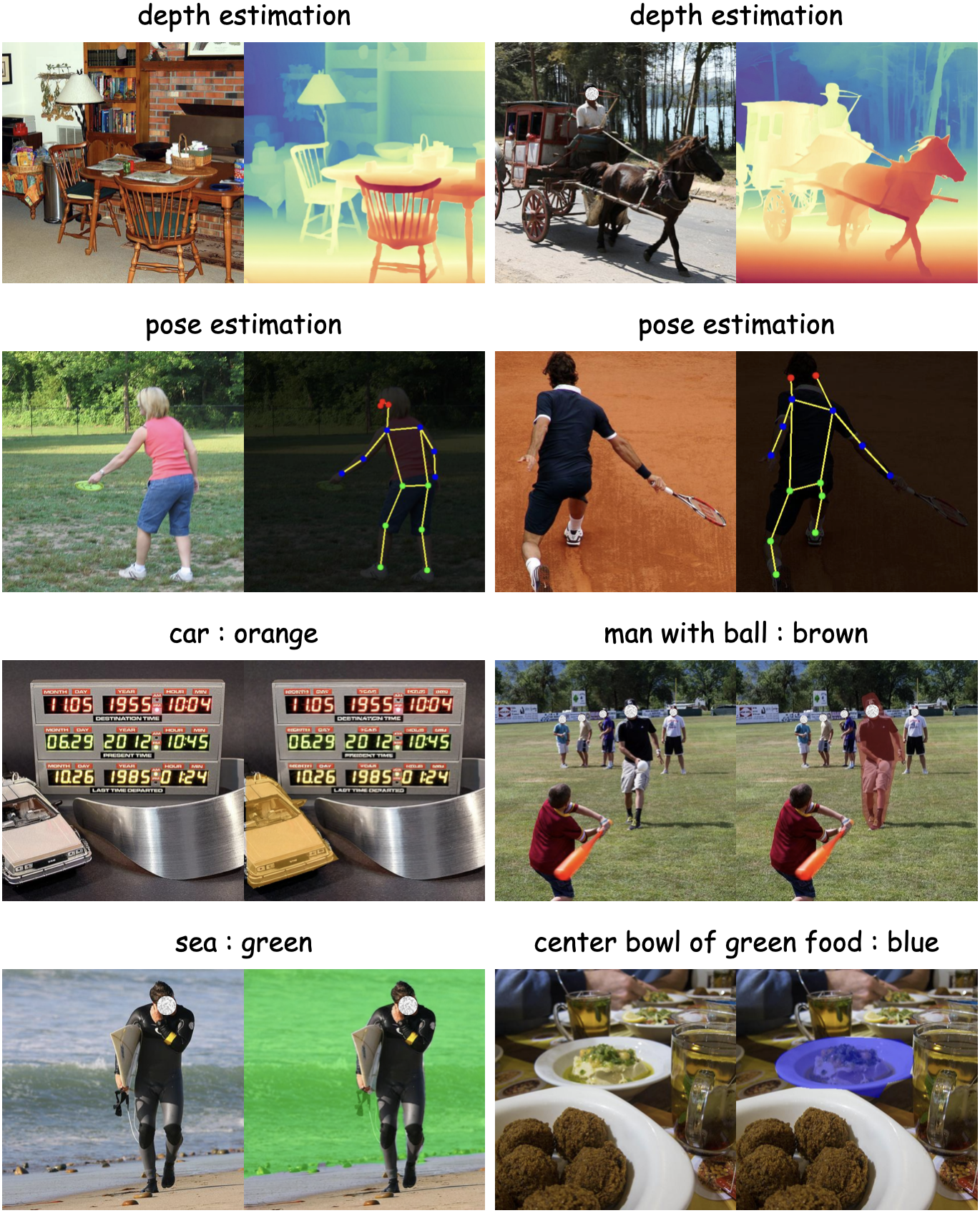}
    \vspace{-1.5ex}
    \caption{Qualitative examples of auxiliary tasks, including depth estimation, pose estimation, and referring segmentation.}
    \vspace{-3ex}
    \label{fig:auxiliary-task}
\end{figure}

Fig.~\ref{fig:auxiliary-task} illustrates that our \model~also supports computer vision applications in auxiliary tasks, such as depth estimation and pose estimation. Regarding referring segmentation, we can recognize the precise object (\textit{e.g.}, \textit{car} and \textit{man with ball}) or even split out the background (\textit{e.g.}, \textit{sea}). This finding also presents the potential of treating our model to unify both visual understanding and generation.

\section{Conclusion}
In this paper, we introduce \model, a generalist diffusion model that unifies a diverse set of image generation tasks within a single framework. We conduct a thorough study and ablate key modeling and data choices, adopting a minimalist model architecture design and introducing a three-stage training pipeline. Our results demonstrate that a single model can effectively handle all tasks without compromising the core text-to-image generation performance. For future work, we plan to incorporate additional vision perception tasks into the framework and further scale up the model to train an even more powerful generalist model.

\vspace{1mm}
\noindent\textbf{Acknowledgement.} We thank Jiasen Lu, Yifang Jiang, and many others for their invaluable help and feedback.

\small
\bibliographystyle{ieeenat_fullname}
\bibliography{main}

\begin{thebibliography}{73}
\providecommand{\natexlab}[1]{#1}
\providecommand{\url}[1]{\texttt{#1}}
\expandafter\ifx\csname urlstyle\endcsname\relax
  \providecommand{\doi}[1]{doi: #1}\else
  \providecommand{\doi}{doi: \begingroup \urlstyle{rm}\Url}\fi

\bibitem[Albergo and Vanden-Eijnden(2023)]{albergo2023flow}
Michael Albergo and Eric Vanden-Eijnden.
\newblock {Building Normalizing Flows with Stochastic Interpolants}.
\newblock In \emph{International Conference on Learning Representations (ICLR)}, 2023.

\bibitem[Astruc et~al.(2024)Astruc, Dufour, Siglidis, Aronssohn, Bouia, Fu, Loiseau, Nguyen, Raude, Vincent, XU, Zhou, and Landrieu]{astruc2024osv-5m}
Guillaume Astruc, Nicolas Dufour, Ioannis Siglidis, Constantin Aronssohn, Nacim Bouia, Stephanie Fu, Romain Loiseau, Van~Nguyen Nguyen, Charles Raude, Elliot Vincent, Lintao XU, Hongyu Zhou, and Loic Landrieu.
\newblock {OpenStreetView-5M: The Many Roads to Global Visual Geolocation}.
\newblock In \emph{Conference on Computer Vision and Pattern Recognition (CVPR)}, 2024.

\bibitem[Bao et~al.(2024)Bao, Nie, Xue, Cao, Li, Su, and Zhu]{bao2023all}
Fan Bao, Shen Nie, Kaiwen Xue, Yue Cao, Chongxuan Li, Hang Su, and Jun Zhu.
\newblock {All are Worth Words: A ViT Backbone for Diffusion Models}.
\newblock In \emph{Conference on Computer Vision and Pattern Recognition (CVPR)}, 2024.

\bibitem[Brooks et~al.(2023)Brooks, Holynski, and Efros]{brooks2023ins-p2p}
Tim Brooks, Aleksander Holynski, and Alexei Efros.
\newblock {InstructPix2Pix: Learning to Follow Image Editing Instructions}.
\newblock In \emph{Conference on Computer Vision and Pattern Recognition (CVPR)}, 2023.

\bibitem[Chen et~al.(2024)Chen, Yu, Ge, Yao, Xie, Wu, Wang, Kwok, Luo, Lu, and Li]{chen2023pixart}
Junsong Chen, Jincheng Yu, Chongjian Ge, Lewei Yao, Enze Xie, Yue Wu, Zhongdao Wang, James Kwok, Ping Luo, Huchuan Lu, and Zhenguo Li.
\newblock {PixArt-alpha: Fast Training of Diffusion Transformer for Photorealistic Text-to-Image Synthesis}.
\newblock In \emph{International Conference on Learning Representations (ICLR)}, 2024.

\bibitem[Chen et~al.(2018)Chen, Rubanova, Bettencourt, and Duvenaud]{chen2018ode}
Ricky Chen, Yulia Rubanova, Jesse Bettencourt, and David Duvenaud.
\newblock {Neural Ordinary Differential Equations}.
\newblock In \emph{Conference on Neural Information Processing Systems (NeurIPS)}, 2018.

\bibitem[Cho et~al.(2024)Cho, Hu, Garg, Anderson, Krishna, Baldridge, Bansal, Pont-Tuset, and Wang]{cho2024dsg}
Jaemin Cho, Yushi Hu, Roopal Garg, Peter Anderson, Ranjay Krishna, Jason Baldridge, Mohit Bansal, Jordi Pont-Tuset, and Su Wang.
\newblock {Davidsonian Scene Graph: Improving Reliability in Fine-grained Evaluation for Text-to-Image Generation}.
\newblock In \emph{International Conference on Learning Representations (ICLR)}, 2024.

\bibitem[Dhariwal and Nichol(2021)]{dhariwal2021diffusion}
Prafulla Dhariwal and Alex Nichol.
\newblock {Diffusion Models Beat GANs on Image Synthesis}.
\newblock In \emph{Conference on Neural Information Processing Systems (NeurIPS)}, 2021.

\bibitem[Egan and Redden(2024)]{egan2024dalle3-1m}
Ben Egan and Alex Redden.
\newblock {Dalle3 1 Million+ High Quality Captions}, 2024.

\bibitem[Esser et~al.(2024)Esser, Kulal, Blattmann, Entezari, Müller, Saini, Levi, Lorenz, Sauer, Boesel, Podell, Dockhorn, English, Lacey, Goodwin, Marek, and Rombach]{esser2024sd3}
Patrick Esser, Sumith Kulal, Andreas Blattmann, Rahim Entezari, Jonas Müller, Harry Saini, Yam Levi, Dominik Lorenz, Axel Sauer, Frederic Boesel, Dustin Podell, Tim Dockhorn, Zion English, Kyle Lacey, Alex Goodwin, Yannik Marek, and Robin Rombach.
\newblock {Scaling Rectified Flow Transformers for High-Resolution Image Synthesis}.
\newblock In \emph{arXiv:2403.03206}, 2024.

\bibitem[Fu et~al.(2024)Fu, Hu, Du, Wang, Yang, and Gan]{fu2024mgie}
Tsu-Jui Fu, Wenze Hu, Xianzhi Du, William~Yang Wang, Yinfei Yang, and Zhe Gan.
\newblock {Guiding Instruction-based Image Editing via Multimodal Large Language Models}.
\newblock In \emph{International Conference on Learning Representations (ICLR)}, 2024.

\bibitem[Gal et~al.(2024)Gal, Lichter, Richardson, Patashnik, Bermano, Chechik, and Cohen-Or]{gal2024unsplash-50}
Rinon Gal, Or Lichter, Elad Richardson, Or Patashnik, Amit Bermano, Gal Chechik, and Daniel Cohen-Or.
\newblock {LCM-Lookahead for Encoder-based Text-to-Image Personalization}.
\newblock In \emph{European Conference on Computer Vision (ECCV)}, 2024.

\bibitem[Ge et~al.(2024)Ge, Zhao, Li, Ge, and Shan]{ge2024seed-edit}
Yuying Ge, Sijie Zhao, Chen Li, Yixiao Ge, and Ying Shan.
\newblock {SEED-Data-Edit Technical Report: A Hybrid Dataset for Instructional Image Editing}.
\newblock In \emph{arXiv:2405.04007}, 2024.

\bibitem[Geiger et~al.(2012)Geiger, Lenz, and Urtasun]{geiger2012kitti}
Andreas Geiger, Philip Lenz, and Raquel Urtasun.
\newblock {Are we ready for Autonomous Driving? The KITTI Vision Benchmark Suite}.
\newblock In \emph{Conference on Computer Vision and Pattern Recognition (CVPR)}, 2012.

\bibitem[Ghosh et~al.(2023)Ghosh, Hajishirzi, and Schmidt]{ghosh2023gen-eval}
Dhruba Ghosh, Hanna Hajishirzi, and Ludwig Schmidt.
\newblock {GenEval: An Object-Focused Framework for Evaluating Text-to-Image Alignment}.
\newblock In \emph{Conference on Neural Information Processing Systems (NeurIPS)}, 2023.

\bibitem[Google(2024)]{baldridge2024imagen}
Imagen~Team Google.
\newblock {Imagen 3}.
\newblock In \emph{arXiv:2408.07009}, 2024.

\bibitem[Guo et~al.(2024)Guo, Wu, Chen, Chen, Zhang, and He]{guo2024pulid}
Zinan Guo, Yanze Wu, Zhuowei Chen, Lang Chen, Peng Zhang, and Qian He.
\newblock {PuLID: Pure and Lightning ID Customization via Contrastive Alignment}.
\newblock In \emph{Conference on Neural Information Processing Systems (NeurIPS)}, 2024.

\bibitem[Han et~al.(2024)Han, Mao, Jiang, Pan, and Zhang]{han2024style-booth}
Zhen Han, Chaojie Mao, Zeyinzi Jiang, Yulin Pan, and Jingfeng Zhang.
\newblock {StyleBooth: Image Style Editing with Multimodal Instruction}.
\newblock In \emph{arXiv:2404.12154}, 2024.

\bibitem[Hatamizadeh et~al.(2024)Hatamizadeh, Song, Liu, Kautz, and Vahdat]{hatamizadeh2024diffit}
Ali Hatamizadeh, Jiaming Song, Guilin Liu, Jan Kautz, and Arash Vahdat.
\newblock {DiffiT: Diffusion Vision Transformers for Image Generation}.
\newblock In \emph{European Conference on Computer Vision (ECCV)}, 2024.

\bibitem[Hertz et~al.(2023)Hertz, Mokady, Tenenbaum, Aberman, Pritch, and Cohen-Or]{hertz2023prompt-to-prompt}
Amir Hertz, Ron Mokady, Jay Tenenbaum, Kfir Aberman, Yael Pritch, and Daniel Cohen-Or.
\newblock {Prompt-to-Prompt Image Editing with Cross Attention Control}.
\newblock In \emph{International Conference on Learning Representations (ICLR)}, 2023.

\bibitem[Hessel et~al.(2021)Hessel, Holtzman, Forbes, Bras, and Choi]{hessel2021clip-score}
Jack Hessel, Ari Holtzman, Maxwell Forbes, Ronan~Le Bras, and Yejin Choi.
\newblock {CLIPScore: A Reference-free Evaluation Metric for Image Captioning}.
\newblock In \emph{Conference on Empirical Methods in Natural Language Processing (EMNLP)}, 2021.

\bibitem[Ho and Salimans(2022)]{ho2022cfg}
Jonathan Ho and Tim Salimans.
\newblock {Classifier-Free Diffusion Guidance}.
\newblock In \emph{Conference on Neural Information Processing Systems (NeurIPS)}, 2022.

\bibitem[Ho et~al.(2020)Ho, Jain, and Abbeel]{ho2020ddpm}
Jonathan Ho, Ajay Jain, and Pieter Abbeel.
\newblock {Denoising Diffusion Probabilistic Models}.
\newblock In \emph{Conference on Neural Information Processing Systems (NeurIPS)}, 2020.

\bibitem[Huang et~al.(2023)Huang, Sun, Xie, Li, and Liu]{huang2023comp-bench}
Kaiyi Huang, Kaiyue Sun, Enze Xie, Zhenguo Li, and Xihui Liu.
\newblock {T2I-CompBench: A Comprehensive Benchmark for Open-world Compositional Text-to-image Generation}.
\newblock In \emph{Conference on Neural Information Processing Systems (NeurIPS)}, 2023.

\bibitem[Huang et~al.(2020)Huang, Wang, Tai, Liu, Shen, Li, Li, and Huang]{huang2020curricular-face}
Yuge Huang, Yuhan Wang, Ying Tai, Xiaoming Liu, Pengcheng Shen, Shaoxin Li, Jilin Li, and Feiyue Huang.
\newblock {CurricularFace: Adaptive Curriculum Learning Loss for Deep Face Recognition}.
\newblock In \emph{Conference on Computer Vision and Pattern Recognition (CVPR)}, 2020.

\bibitem[Labs(2024)]{flux1}
Black~Forest Labs.
\newblock {FLUX.1 [dev]}, 2024.

\bibitem[Lai et~al.(2024)Lai, Zhang, Zhang, Wu, Bai, Timofeev, Du, Gan, Shan, Chuah, Yang, and Cao]{lai2024ve-clip}
Zhengfeng Lai, Haotian Zhang, Bowen Zhang, Wentao Wu, Haoping Bai, Aleksei Timofeev, Xianzhi Du, Zhe Gan, Jiulong Shan, Chen-Nee Chuah, Yinfei Yang, and Meng Cao.
\newblock {VeCLIP: Improving CLIP Training via Visual-enriched Captions}.
\newblock In \emph{European Conference on Computer Vision (ECCV)}, 2024.

\bibitem[Le et~al.(2024)Le, Pham, Lee, Clark, Kembhavi, Mandt, Krishna, and Lu]{le2024one-diff}
Duong Le, Tuan Pham, Sangho Lee, Christopher Clark, Aniruddha Kembhavi, Stephan Mandt, Ranjay Krishna, and Jiasen Lu.
\newblock {One Diffusion to Generate Them All}.
\newblock In \emph{arXiv:2411.16318}, 2024.

\bibitem[Li et~al.(2023{\natexlab{a}})Li, Li, and Hoi]{li2023blip}
Dongxu Li, Junnan Li, and Steven Hoi.
\newblock {BLIP-Diffusion: Pre-trained Subject Representation for Controllable Text-to-Image Generation and Editing}.
\newblock In \emph{Conference on Neural Information Processing Systems (NeurIPS)}, 2023{\natexlab{a}}.

\bibitem[Li et~al.(2023{\natexlab{b}})Li, Liu, Wu, Mu, Yang, Gao, Li, and Lee]{li2023gligen}
Yuheng Li, Haotian Liu, Qingyang Wu, Fangzhou Mu, Jianwei Yang, Jianfeng Gao, Chunyuan Li, and Yong~Jae Lee.
\newblock {GLIGEN: Open-Set Grounded Text-to-Image Generation}.
\newblock In \emph{Conference on Computer Vision and Pattern Recognition (CVPR)}, 2023{\natexlab{b}}.

\bibitem[Li et~al.(2023{\natexlab{c}})Li, Cao, Wang, Qi, Cheng, and Shan]{li2023photo-maker}
Zhen Li, Mingdeng Cao, Xintao Wang, Zhongang Qi, Ming-Ming Cheng, and Ying Shan.
\newblock {PhotoMaker: Customizing Realistic Human Photos via Stacked ID Embedding}.
\newblock In \emph{arXiv:2312.04461}, 2023{\natexlab{c}}.

\bibitem[Lin et~al.(2014)Lin, Maire, Belongie, Bourdev, Girshick, Hays, Perona, Ramanan, Zitnick, and Dollar]{lin2014coco}
Tsung-Yi Lin, Michael Maire, Serge Belongie, Lubomir Bourdev, Ross Girshick, James Hays, Pietro Perona, Deva Ramanan, Lawrence Zitnick, and Piotr Dollar.
\newblock {Microsoft COCO: Common Objects in Context}.
\newblock In \emph{European Conference on Computer Vision (ECCV)}, 2014.

\bibitem[Lipman et~al.(2023)Lipman, Chen, Ben-Hamu, Nickel, and Le]{lipman2023flow-matching}
Yaron Lipman, Ricky Chen, Heli Ben-Hamu, Maximilian Nickel, and Matt Le.
\newblock {Flow Matching for Generative Modeling}.
\newblock In \emph{International Conference on Learning Representations (ICLR)}, 2023.

\bibitem[Liu et~al.(2023)Liu, Gong, and Liu]{liu2023flow}
Xingchao Liu, Chengyue Gong, and Qiang Liu.
\newblock {Flow Straight and Fast: Learning to Generate and Transfer Data with Rectified Flow}.
\newblock In \emph{International Conference on Learning Representations (ICLR)}, 2023.

\bibitem[Lu et~al.(2023)Lu, Clark, Zellers, Mottaghi, and Kembhavi]{lu2023unified}
Jiasen Lu, Christopher Clark, Rowan Zellers, Roozbeh Mottaghi, and Aniruddha Kembhavi.
\newblock {Unified-IO: A Unified Model for Vision, Language, and Multi-Modal Tasks}.
\newblock In \emph{International Conference on Learning Representations (ICLR)}, 2023.

\bibitem[Lu et~al.(2024)Lu, Clark, Lee, Zhang, Khosla, Marten, Hoiem, and Kembhavi]{lu2024unified}
Jiasen Lu, Christopher Clark, Sangho Lee, Zichen Zhang, Savya Khosla, Ryan Marten, Derek Hoiem, and Aniruddha Kembhavi.
\newblock {Unified-IO 2: Scaling Autoregressive Multimodal Models with Vision, Language, Audio, and Action}.
\newblock In \emph{Conference on Computer Vision and Pattern Recognition (CVPR)}, 2024.

\bibitem[Ma et~al.(2024)Ma, Goldstein, Albergo, Boffi, Vanden-Eijnden, and Xie]{ma2024sit}
Nanye Ma, Mark Goldstein, Michael Albergo, Nicholas Boffi, Eric Vanden-Eijnden, and Saining Xie.
\newblock {SiT: Exploring Flow and Diffusion-based Generative Models with Scalable Interpolant Transformers}.
\newblock In \emph{European Conference on Computer Vision (ECCV)}, 2024.

\bibitem[Mao et~al.(2025)Mao, Zhang, Pan, Jiang, Han, Liu, and Zhou]{mao2025ace++}
Chaojie Mao, Jingfeng Zhang, Yulin Pan, Zeyinzi Jiang, Zhen Han, Yu Liu, and Jingren Zhou.
\newblock {ACE++: Instruction-Based Image Creation and Editing via Context-Aware Content Filling}.
\newblock In \emph{arXiv:2501.02487}, 2025.

\bibitem[Mou et~al.(2024)Mou, Wang, Xie, Wu, Zhang, Qi, Shan, and Qie]{mou2024t2i}
Chong Mou, Xintao Wang, Liangbin Xie, Yanze Wu, Jian Zhang, Zhongang Qi, Ying Shan, and Xiaohu Qie.
\newblock {T2I-Adapter: Learning Adapters to Dig out More Controllable Ability for Text-to-Image Diffusion Models}.
\newblock In \emph{Association for the Advancement of Artificial Intelligence (AAAI)}, 2024.

\bibitem[Nichol et~al.(2022)Nichol, Dhariwal, Ramesh, Shyam, Mishkin, McGrew, Sutskever, and Chen]{nichol2021glide}
Alex Nichol, Prafulla Dhariwal, Aditya Ramesh, Pranav Shyam, Pamela Mishkin, Bob McGrew, Ilya Sutskever, and Mark Chen.
\newblock {GLIDE: Towards Photorealistic Image Generation and Editing with Text-Guided Diffusion Models}.
\newblock In \emph{International Conference on Machine Learning (ICML)}, 2022.

\bibitem[Ordonez et~al.(2021)Ordonez, Kulkarni, and Berg]{ordonez2021sbu}
Vicente Ordonez, Girish Kulkarni, and Tamara Berg.
\newblock {Im2Text: Describing Images Using 1 Million Captioned Photographs}.
\newblock In \emph{Conference on Neural Information Processing Systems (NeurIPS)}, 2021.

\bibitem[Peebles and Xie(2022)]{peebles2022dit}
William Peebles and Saining Xie.
\newblock {Scalable Diffusion Models with Transformers}.
\newblock In \emph{arXiv:2212.09748}, 2022.

\bibitem[Podell et~al.(2023)Podell, English, Lacey, Blattmann, Dockhorn, Müller, Penna, and Rombach]{podell2023sdxl}
Dustin Podell, Zion English, Kyle Lacey, Andreas Blattmann, Tim Dockhorn, Jonas Müller, Joe Penna, and Robin Rombach.
\newblock {SDXL: Improving Latent Diffusion Models for High-Resolution Image Synthesis}.
\newblock In \emph{arXiv:2307.01952}, 2023.

\bibitem[Qin et~al.(2023)Qin, Zhang, Yu, Feng, Yang, Zhou, Wang, Niebles, Xiong, Savarese, Ermon, Fu, and Xu]{qin2023unicontrol}
Can Qin, Shu Zhang, Ning Yu, Yihao Feng, Xinyi Yang, Yingbo Zhou, Huan Wang, Juan~Carlos Niebles, Caiming Xiong, Silvio Savarese, Stefano Ermon, Yun Fu, and Ran Xu.
\newblock {UniControl: A Unified Diffusion Model for Controllable Visual Generation In the Wild}.
\newblock In \emph{Conference on Neural Information Processing Systems (NeurIPS)}, 2023.

\bibitem[Ramesh et~al.(2022)Ramesh, Dhariwal, Nichol, Chu, and Chen]{ramesh2022hierarchical}
Aditya Ramesh, Prafulla Dhariwal, Alex Nichol, Casey Chu, and Mark Chen.
\newblock {Hierarchical Text-Conditional Image Generation with CLIP Latents}.
\newblock In \emph{arXiv:2204.06125}, 2022.

\bibitem[Roberts et~al.(2021)Roberts, Ramapuram, Ranjan, Kumar, Bautista, Paczan, Webb, and Susskind]{roberts2021hypersim}
Mike Roberts, Jason Ramapuram, Anurag Ranjan, Atulit Kumar, Miguel~Angel Bautista, Nathan Paczan, Russ Webb, and Joshua Susskind.
\newblock {Hypersim: A Photorealistic Synthetic Dataset for Holistic Indoor Scene Understanding}.
\newblock In \emph{International Conference on Computer Vision (ICCV)}, 2021.

\bibitem[Rombach et~al.(2022)Rombach, Blattmann, Lorenz, Esser, and Ommer]{rombach2022ldm}
Robin Rombach, Andreas Blattmann, Dominik Lorenz, Patrick Esser, and Björn Ommer.
\newblock {High-Resolution Image Synthesis with Latent Diffusion Models}.
\newblock In \emph{Conference on Computer Vision and Pattern Recognition (CVPR)}, 2022.

\bibitem[Ronneberger et~al.(2015)Ronneberger, Fischer, and Brox]{ronneberger2015u-net}
Olaf Ronneberger, Philipp Fischer, and Thomas Brox.
\newblock {U-Net: Convolutional Networks for Biomedical Image Segmentation}.
\newblock In \emph{International Conference on Medical Image Computing and Computer-Assisted Intervention (MICCAI)}, 2015.

\bibitem[Shazeer and Stern(2018)]{shazeer2018adafactor}
Noam Shazeer and Mitchell Stern.
\newblock {Adafactor: Adaptive Learning Rates with Sublinear Memory Cost}.
\newblock In \emph{International Conference on Machine Learning (ICML)}, 2018.

\bibitem[Sheynin et~al.(2023)Sheynin, Polyak, Singer, Kirstain, Zohar, Ashual, Parikh, and Taigman]{sheynin2023emu-edit}
Shelly Sheynin, Adam Polyak, Uriel Singer, Yuval Kirstain, Amit Zohar, Oron Ashual, Devi Parikh, and Yaniv Taigman.
\newblock {Emu Edit: Precise Image Editing via Recognition and Generation Tasks}.
\newblock In \emph{arXiv:2311.10089}, 2023.

\bibitem[Song et~al.(2021)Song, Meng, and Ermon]{song2021ddim}
Jiaming Song, Chenlin Meng, and Stefano Ermon.
\newblock {Denoising Diffusion Implicit Models}.
\newblock In \emph{International Conference on Learning Representations (ICLR)}, 2021.

\bibitem[Song and Ermon(2019)]{song2019generative}
Yang Song and Stefano Ermon.
\newblock {Generative modeling by estimating gradients of the data distribution}.
\newblock In \emph{Conference on Neural Information Processing Systems (NeurIPS)}, 2019.

\bibitem[Sun et~al.(2023)Sun, Pan, Ge, Li, Duan, Wu, Zhang, Zhou, Qin, Wang, Dai, Qiao, Wang, and Li]{sun2023jdb}
Keqiang Sun, Junting Pan, Yuying Ge, Hao Li, Haodong Duan, Xiaoshi Wu, Renrui Zhang, Aojun Zhou, Zipeng Qin, Yi Wang, Jifeng Dai, Yu Qiao, Limin Wang, and Hongsheng Li.
\newblock {JourneyDB: A Benchmark for Generative Image Understanding}.
\newblock In \emph{Conference on Neural Information Processing Systems (NeurIPS)}, 2023.

\bibitem[Team(2024)]{team2024chameleon}
Chameleon Team.
\newblock {Chameleon: Mixed-Modal Early-Fusion Foundation Models}.
\newblock In \emph{arXiv:2405.09818}, 2024.

\bibitem[Tong et~al.(2023)Tong, Fatras, Malkin, Huguet, Zhang, Rector-Brooks, Wolf, and Bengio]{tong2023improving}
Alexander Tong, Kilian Fatras, Nikolay Malkin, Guillaume Huguet, Yanlei Zhang, Jarrid Rector-Brooks, Guy Wolf, and Yoshua Bengio.
\newblock {Improving and generalizing flow-based generative models with minibatch optimal transport}.
\newblock In \emph{Transactions on Machine Learning Research (TMLR)}, 2023.

\bibitem[Wang et~al.(2024)Wang, Bai, Wang, Qin, Chen, Li, Tang, and Hu]{wang2024instant-id}
Qixun Wang, Xu Bai, Haofan Wang, Zekui Qin, Anthony Chen, Huaxia Li, Xu Tang, and Yao Hu.
\newblock {InstantID: Zero-shot Identity-Preserving Generation in Seconds}.
\newblock In \emph{arXiv:2401.07519}, 2024.

\bibitem[Wei et~al.(2024)Wei, Xiong, Ren, Du, Zhang, and Chen]{wei2024omni-edit}
Cong Wei, Zheyang Xiong, Weiming Ren, Xinrun Du, Ge Zhang, and Wenhu Chen.
\newblock {OmniEdit: Building Image Editing Generalist Models Through Specialist Supervision}.
\newblock In \emph{arXiv:2411.07199}, 2024.

\bibitem[Wu et~al.(2020)Wu, Lin, Cohen, Bui, and Maji]{wu2020phrase-cut}
Chenyun Wu, Zhe Lin, Scott Cohen, Trung Bui, and Subhransu Maji.
\newblock {PhraseCut: Language-based Image Segmentation in the Wild}.
\newblock In \emph{Conference on Computer Vision and Pattern Recognition (CVPR)}, 2020.

\bibitem[Wu et~al.(2023)Wu, Hao, Sun, Chen, Zhu, Zhao, and Li]{wu2023hpsv2}
Xiaoshi Wu, Yiming Hao, Keqiang Sun, Yixiong Chen, Feng Zhu, Rui Zhao, and Hongsheng Li.
\newblock {Human Preference Score v2: A Solid Benchmark for Evaluating Human Preferences of Text-to-Image Synthesis}.
\newblock In \emph{arXiv:2306.09341}, 2023.

\bibitem[Xiao et~al.(2024)Xiao, Wang, Zhou, Yuan, Xing, Yan, Li, Wang, Huang, and Liu]{xiao2024omni-gen}
Shitao Xiao, Yueze Wang, Junjie Zhou, Huaying Yuan, Xingrun Xing, Ruiran Yan, Chaofan Li, Shuting Wang, Tiejun Huang, and Zheng Liu.
\newblock {OmniGen: Unified Image Generation}.
\newblock In \emph{arXiv:2409.11340}, 2024.

\bibitem[Xie et~al.(2025)Xie, Mao, Bai, Zhang, Wang, Lin, Gu, Chen, Yang, and Shou]{xie2024show}
Jinheng Xie, Weijia Mao, Zechen Bai, David~Junhao Zhang, Weihao Wang, Kevin~Qinghong Lin, Yuchao Gu, Zhijie Chen, Zhenheng Yang, and Mike~Zheng Shou.
\newblock {Show-o: One Single Transformer to Unify Multimodal Understanding and Generation}.
\newblock In \emph{International Conference on Learning Representations (ICLR)}, 2025.

\bibitem[Xu et~al.(2023)Xu, Wang, Zhang, Wang, and Shi]{xu2023versatile}
Xingqian Xu, Zhangyang Wang, Eric Zhang, Kai Wang, and Humphrey Shi.
\newblock {Versatile Diffusion: Text, Images and Variations All in One Diffusion Model}.
\newblock In \emph{International Conference on Computer Vision (ICCV)}, 2023.

\bibitem[Yang et~al.(2022)Yang, Gan, Wang, Hu, Ahmed, Liu, Lu, and Wang]{yang2022unitab}
Zhengyuan Yang, Zhe Gan, Jianfeng Wang, Xiaowei Hu, Faisal Ahmed, Zicheng Liu, Yumao Lu, and Lijuan Wang.
\newblock {UniTAB: Unifying Text and Box Outputs for Grounded Vision-Language Modeling}.
\newblock In \emph{European Conference on Computer Vision (ECCV)}, 2022.

\bibitem[Yang et~al.(2023)Yang, Wang, Gan, Li, Lin, Wu, Duan, Liu, Liu, Zeng, and Wang]{yang2023reco}
Zhengyuan Yang, Jianfeng Wang, Zhe Gan, Linjie Li, Kevin Lin, Chenfei Wu, Nan Duan, Zicheng Liu, Ce Liu, Michael Zeng, and Lijuan Wang.
\newblock {ReCo: Region-Controlled Text-to-Image Generation}.
\newblock In \emph{Conference on Computer Vision and Pattern Recognition (CVPR)}, 2023.

\bibitem[Ye et~al.(2024)Ye, Zhang, Liu, Han, and Yang]{ye2023ip}
Hu Ye, Jun Zhang, Sibo Liu, Xiao Han, and Wei Yang.
\newblock {IP-Adapter: Text Compatible Image Prompt Adapter for Text-to-Image Diffusion Models}.
\newblock In \emph{arXiv:2308.06721}, 2024.

\bibitem[Young et~al.(2014)Young, Lai, Hodosh, and Hockenmaier]{young2014flickr30k}
Peter Young, Alice Lai, Micah Hodosh, and Julia Hockenmaier.
\newblock {From Image Descriptions to Visual Denotations: New Similarity Metrics for Semantic Inference over Event Descriptions}.
\newblock In \emph{Transactions of the Association for Computational Linguistics (TACL)}, 2014.

\bibitem[Yu et~al.(2016)Yu, Poirson, Yang, Berg, and Berg]{yu2016ref-coco}
Licheng Yu, Patrick Poirson, Shan Yang, Alexander Berg, and Tamara Berg.
\newblock {Modeling Context in Referring Expressions}.
\newblock In \emph{European Conference on Computer Vision (ECCV)}, 2016.

\bibitem[Zhang et~al.(2023{\natexlab{a}})Zhang, Mo, Chen, Sun, and Su]{zhang2023magic-brush}
Kai Zhang, Lingbo Mo, Wenhu Chen, Huan Sun, and Yu Su.
\newblock {MagicBrush: A Manually Annotated Dataset for Instruction-Guided Image Editing}.
\newblock In \emph{Conference on Neural Information Processing Systems (NeurIPS)}, 2023{\natexlab{a}}.

\bibitem[Zhang et~al.(2023{\natexlab{b}})Zhang, Rao, and Agrawala]{zhang2023adding}
Lvmin Zhang, Anyi Rao, and Maneesh Agrawala.
\newblock {Adding conditional control to text-to-image diffusion models}.
\newblock In \emph{International Conference on Computer Vision (ICCV)}, 2023{\natexlab{b}}.

\bibitem[Zhao et~al.(2024)Zhao, Ma, Chen, Si, Wu, An, Yu, Zhang, Li, and Chang]{zhao2024ultra-edit}
Haozhe Zhao, Xiaojian Ma, Liang Chen, Shuzheng Si, Rujie Wu, Kaikai An, Peiyu Yu, Minjia Zhang, Qing Li, and Baobao Chang.
\newblock {UltraEdit: Instruction-based Fine-Grained Image Editing at Scale}.
\newblock In \emph{Conference on Neural Information Processing Systems (NeurIPS)}, 2024.

\bibitem[Zhao et~al.(2023)Zhao, Chen, Chen, Bao, Hao, Yuan, and Wong]{zhao2023uni}
Shihao Zhao, Dongdong Chen, Yen-Chun Chen, Jianmin Bao, Shaozhe Hao, Lu Yuan, and Kwan-Yee Wong.
\newblock {Uni-ControlNet: All-in-One Control to Text-to-Image Diffusion Models}.
\newblock In \emph{Conference on Neural Information Processing Systems (NeurIPS)}, 2023.

\bibitem[Zhou et~al.(2016)Zhou, Khosla, Lapedriza, Torralba, and Oliva]{zhou2016places365}
Bolei Zhou, Aditya Khosla, Agata Lapedriza, Antonio Torralba, and Aude Oliva.
\newblock {Places: An Image Database for Deep Scene Understanding}.
\newblock In \emph{arXiv:1610.02055}, 2016.

\bibitem[Zhou et~al.(2025)Zhou, Yu, Babu, Tirumala, Yasunaga, Shamis, Kahn, Ma, Zettlemoyer, and Levy]{zhou2024transfusion}
Chunting Zhou, Lili Yu, Arun Babu, Kushal Tirumala, Michihiro Yasunaga, Leonid Shamis, Jacob Kahn, Xuezhe Ma, Luke Zettlemoyer, and Omer Levy.
\newblock {Transfusion: Predict the Next Token and Diffuse Images with One Multi-Modal Model}.
\newblock In \emph{International Conference on Learning Representations (ICLR)}, 2025.

\end{thebibliography}

\end{document}